\documentclass{article}

\usepackage[preprint]{neurips_2024}

\usepackage[utf8]{inputenc} 
\usepackage[T1]{fontenc}    
\usepackage{hyperref}       
\usepackage{url}            
\usepackage{booktabs}       
\usepackage{amsfonts}       
\usepackage{nicefrac}       
\usepackage{microtype}      
\usepackage{xcolor}         
\usepackage[noend]{algpseudocode}

\usepackage{algorithm}
\usepackage{algpseudocode}

\usepackage{graphicx}
\usepackage{amsmath}
\usepackage{natbib}

\title{Mixture of Tokens: Continuous MoE through Cross-Example Aggregation}

\author{Szymon Antoniak  $^{*}$ 
\\
IDEAS NCBR \\
University of Warsaw \\
\And
Michał Krutul $^{*}$ \\
IDEAS NCBR \\
University of Warsaw \\
\And
Maciej Pióro \\
IDEAS NCBR \\
Polish Academy of Sciences \\
\And
Jakub Krajewski \\
IDEAS NCBR \\
University of Warsaw \\
\And
Jan Ludziejewski \\
IDEAS NCBR \\
University of Warsaw \\
\And
Kamil Ciebiera \\
IDEAS NCBR \\
University of Warsaw \\
\And
Krystian Król \\
IDEAS NCBR \\
University of Warsaw \\
\And
Tomasz Odrzygóźdź \\
IDEAS NCBR \\
\And
Marek Cygan \\
University of Warsaw \\
Nomagic \\
\And
Sebastian Jaszczur \thanks{Core contributors. Detailed authors’ contributions are listed in Appendix \ref{sec:contributions}.} \\
IDEAS NCBR \\
University of Warsaw \\
\And
}

\begin{document}

\maketitle

\begin{abstract}
Mixture of Experts (MoE) models based on Transformer architecture are pushing the boundaries of language and vision tasks. The allure of these models lies in their ability to substantially increase the parameter count without a corresponding increase in FLOPs.  Most widely adopted MoE models are discontinuous with respect to their parameters - often referred to as \textit{sparse}. At the same time, existing continuous MoE designs either lag behind their sparse counterparts or are incompatible with autoregressive decoding. Motivated by the observation that the adaptation of fully continuous methods has been an overarching trend in deep learning, we develop Mixture of Tokens (MoT), a simple, continuous architecture that is capable of scaling the number of parameters similarly to sparse MoE models. Unlike conventional methods, MoT assigns mixtures of tokens from different examples to each expert. This architecture is fully compatible with autoregressive training and generation. Our best models not only achieve a $3\times$ increase in training speed over dense Transformer models in language pretraining but also match the performance of state-of-the-art MoE architectures. Additionally, a close connection between MoT and MoE is demonstrated through a novel technique we call \textit{transition tuning}.
\end{abstract}

\section{Introduction}

Transformer-based Large Language Models (LLMs) constitute one of the most active fields in AI, exhibiting human-level performance in a variety of tasks, including translation, language understanding, reasoning, and code generation~\citep{openai2023gpt4, anil2023palm,chowdhery2022palm}. The exorbitant sizes of all state-of-the-art language models are integral to their success, with parameter counts reaching tens or even hundreds of billions. This phenomenon is in line with \citet{kaplan2020scaling} and \citet{hoffmann2022training}, where the latter proposes that the optimal model size grows proportionally to the available compute budget. Given that hardware efficiency has been steadily increasing over the past decade~\citep{compute,compute2}, these findings suggest that scaling will remain a vital component of training increasingly capable models.

\begin{figure} [h!]
    \centering
    \includegraphics[width=0.65\linewidth]{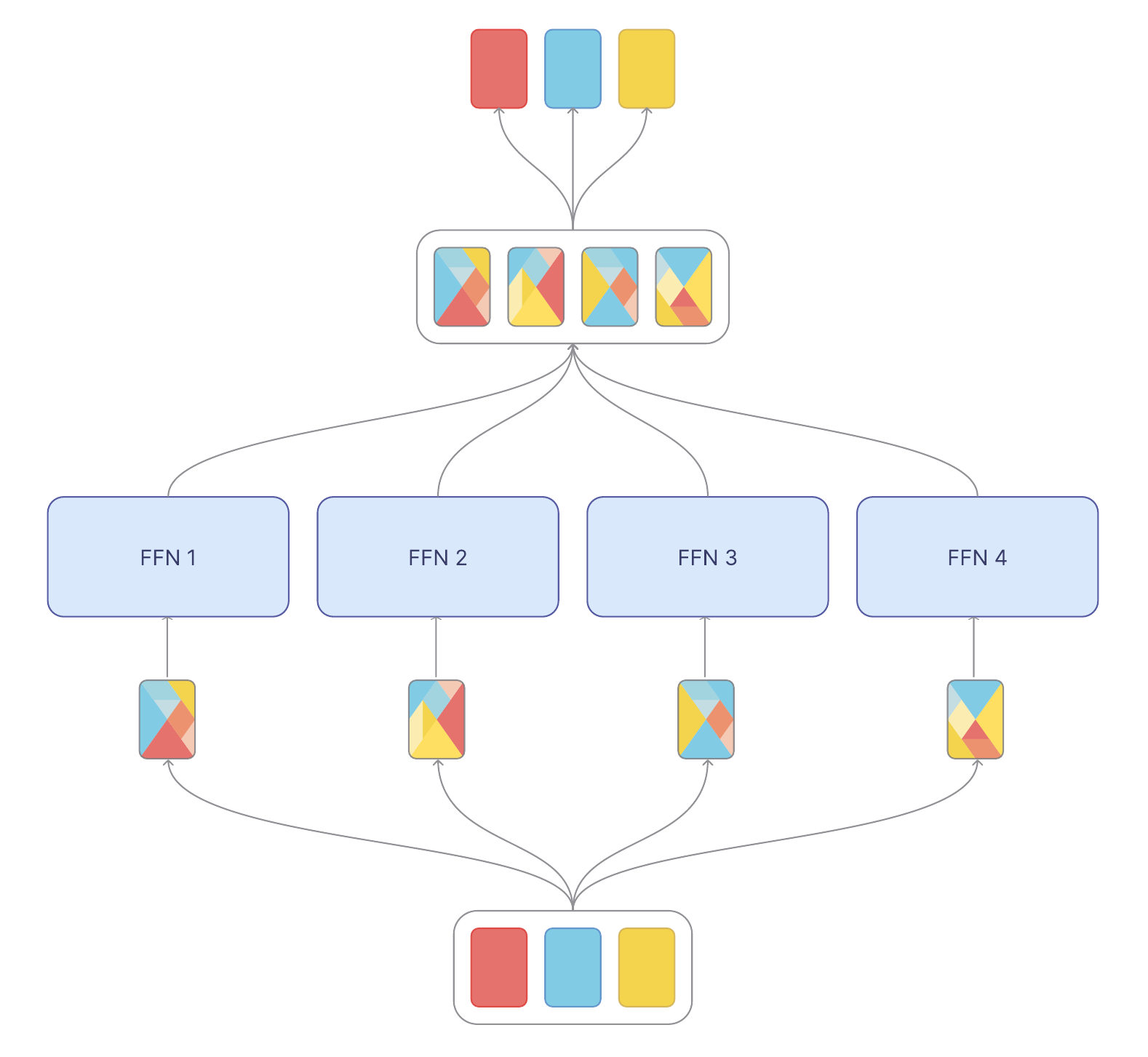}
    \caption{Mixture of Tokens: Each expert receives a unique mixture of tokens in the group. Mixing weights are determined by the controller, which is a fully connected layer (omitted for clarity). For a given token, its update is a linear combination of expert outputs, with the coefficients equal to the token’s original mixing weights for each expert.}
    \label{mot}
\end{figure}

However, model scaling invariably comes at a cost. Larger models execute more Floating Point Operations (FLOPs) per token, which results in both training and inference becoming slower and more expensive~\citep{watts,watts2}. Mixture of Experts~\citep{moe1991} architectures offer an attractive alternative to standard Transformers by drastically increasing the number of parameters. The core idea is to have multiple \textit{experts}, each specializing in a different part of the input space. Currently, state-of-the-art models based on MoE leverage sparsity by activating only a fraction of the parameters for each token~\citep{huggingfacemoe}. This allows the networks to increase the number of parameters by an order of magnitude while keeping the FLOPs per token roughly constant. In this work, we use MoE to signify sparse Mixture of Experts architectures unless explicitly stated otherwise.

The aforementioned sparsity is made possible with a \textit{router}, a small network that selects the best experts for each token. This makes the output of an MoE layer discontinuous with respect to its parameters, as only a subset of the experts is chosen for each token (usually done with a discrete \textit{top-k} operation). The discontinuity and the resulting fluctuations of the router’s decisions were shown to hurt training efficiency~\citep{dai2022stablemoe,collapse} and are hypothesized to be a source of training instability in large MoE models~\citep{limoe,softmoe}. On the other hand, existing continuous MoE architectures involve trade-offs, including inability to scale~\citep{muqeeth2023soft,hazimeh2021dselectk}, or incompatibility with autoregressive decoding~\citep{softmoe}.

This paper introduces Mixture of Tokens, a novel, continuous Transformer architecture closely related to sparse Mixture of Experts. Similarly to MoE, it is capable of supporting large parameter counts without incurring significant costs in FLOPs. The core idea behind our design is to for each expert to process not individual tokens separately but their combined representation.

This technique results in a continuous model that avoids the top-$k$ operation. It requires no additional techniques commonly required in existing MoE designs (both sparse and continuous), such as load balancing losses, calculating solutions to optimization problems, or non-homogeneous training schedules~\citep{hazimeh2021dselectk,jaszczur2021sparse,dai2022stablemoe}. It is capable of scaling the parameter counts akin to sparse MoEs and is compatible with autoregressive language modeling and generation. 

\newpage
Our contributions are the following:

\begin{itemize}
  \item The Mixture of Tokens architecture, a scalable, continuous Mixture of Experts architecture that jointly processes tokens from different examples.
  \item Benchmarking on autoregressive language pretraining against sparse Mixture of Experts architectures, as well as the standard Transformer baseline, over which MoT demonstrates a~$3\times$~speedup.
  \item Analysis of scaling properties of Mixture of Experts models on multiple scales.
  \item Improved performance in low precision training compared to Mixture of Experts.
  \item A novel technique called \textit{transition tuning}, which allows a pre-trained Mixture of Tokens model to use sparse MoE inference once tuned, at a fraction of the pretraining cost.
\end{itemize}

\section{Background and Related Work}

In this section, we provide an overview of the approaches related to our work and discuss the differences between various MoE designs. We will introduce the Mixture of Tokens architecture in Section~\ref{sec:mot} and provide a detailed comparison between MoT and related methods in Section~\ref{comparison}.

\subsection{Large Language Models}
Transformer scaling has been shown to be a critical factor in achieving state-of-the-art results in language and vision tasks~\citep{kaplan2020scaling,zhai2022scaling}, with the largest disclosed parameter counts in dense models reaching hundreds of billions of parameters~\citep{brown2020language,chowdhery2022palm,anil2023palm}. These large models exhibit impressive abilities that are not present in their smaller counterparts~\citep{wei2022emergent}.
\citet{kaplan2020scaling} and \citet{hoffmann2022training} have shown that the final model performance is predictable and correlates directly with the model size and the amount of training data. However, increasing model sizes raises the demand for computational resources during both training and inference~\citep{rae2022scaling}.

\subsection{Mixture of Experts}
Mixture of Experts (MoE) was first introduced by \citet{moe1991} as an ensemble-like neural network comprised of separate sub-networks called experts. The original design used a gating network to select a soft assignment of experts for each input. In the context of Deep Learning, the notion of an MoE \textit{layer} was introduced in \citet{eigen2014learning}. \citet{shazeer2017outrageously} combined a sparse version of MoE layers with an LSTM to train a model with over $100$ billion parameters, unprecedented at the time. The design, similar to state-of-the-art MoE models today, used a small routing network to decide the top-$k$ best experts for each input. By choosing only a subset of the experts, they were able to increase the size of the network while keeping FLOPs per token roughly constant. The Transformer was first combined with MoE layers in \citet{lepikhin2020gshard}, where it replaced the feed-forward layer. The design was further simplified in \citet{fedus2022switch}, which trained a model with $1.6$ trillion parameters with top-$1$ routing. Since then, a number of studies investigated different sparse MoE designs \citep{du2022glam,jiang2024mixtral,zhou2022mixtureofexperts,roller2021hash,lewis2021base}. A comprehensive analysis of scaling properties of sparse MoE architectures can be found in \citet{clark2022unified}.

\begin{figure*}
    \centering
    \medskip
    \includegraphics[width=.8\textwidth]{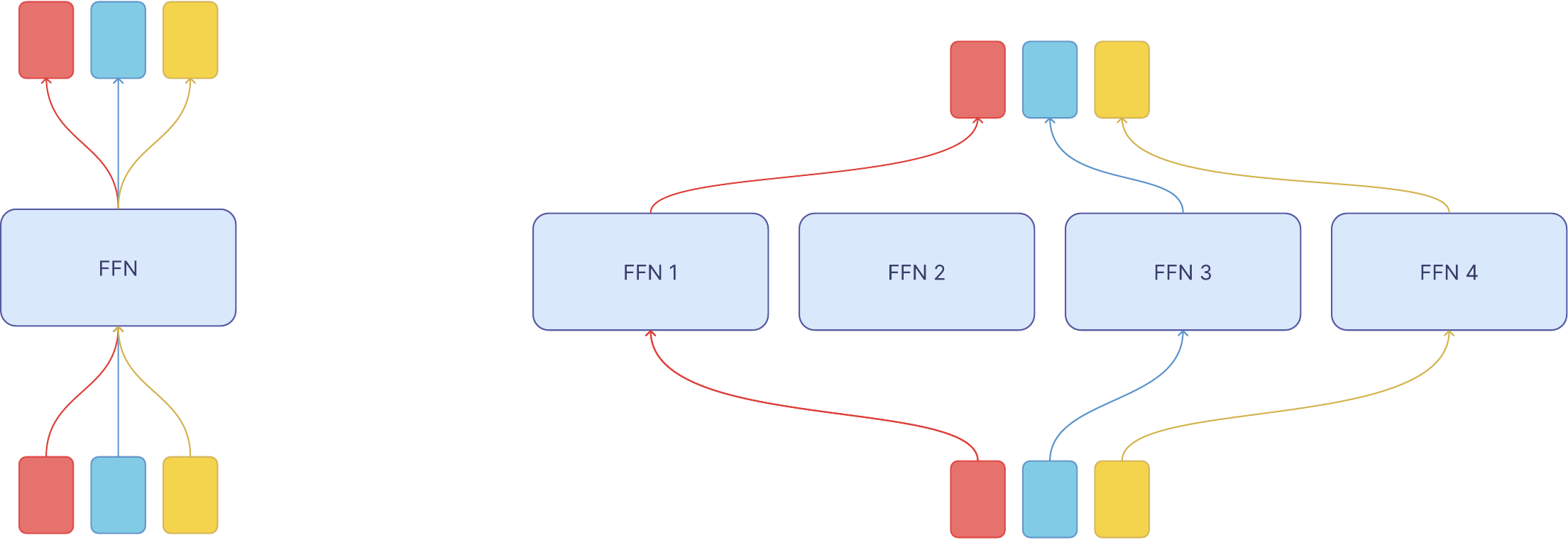}

    \medskip

    \caption{\textbf{(Left)} Diagram of a standard feed-forward layer featured in the Transformer architecture: each token is processed with the same MLP, independent of other tokens. \textbf{(Right)} Diagram of a Token Choice layer, where each token decides which expert to choose. This way, different experts process a different number of tokens. If one expert is chosen by too many tokens, a portion of the tokens are dropped --- they receive no update.
    \label{diag_gran_better}}
\end{figure*}

\subsection{Continuous Mixture of Experts}
 Continuous architectures serve an important role within the field due to their flexible and efficient nature. \citet{hazimeh2021dselectk} were pioneers in introducing them in MoE by presenting continuous techniques for calculating encodings of the choice of an expert.  In another approach, \citet{muqeeth2023soft} proposed a method where they merge experts based on the weights of the router network. In a recent advancement, \citet{softmoe} proposed a continuous variant of MoE for the Vision Transformer, where patches are mixed only within each image.

\subsection{From Hard to Soft Methods}

From the very beginning of the Deep Learning field, we see movements from discrete functions toward continuous ones. The first perceptron \citep{mcculloch1943logical} used "all-or-none" activation, supposedly to align with propositional logic. This was later improved with soft activation functions, enabling gradient descent and multi-layer neural networks. Similarly, soft attention, introduced in \citet{bahdanau2016neural}, enabled RNNs to look at arbitrary input from the past while maintaining the ability to learn the selection with standard gradient descent. This contrasts with hard attention, which requires, for example, reinforcement learning techniques. While hard attention could perform on par with soft attention \citep{hardsoftattention, Zohourianshahzadi_2021}, soft attention, with its simplicity of training, presented better trade-offs and was later used as the basic building block of the Transformer \citep{vaswani2017attention}.

Mixture of Experts, introduced into Deep Learning by \citet{shazeer2017outrageously}, seems like a naturally discrete function --- after all, the expert either processes a given token or it doesn't. However, much like the shift from hard to soft attention, an expert in MoE can "attend" to a mix of tokens, taken as a weighted average. This results in a smooth, continuous model and enables more stable training. 

\section{Mixture of Tokens} \label{sec:mot}

The goal of this work is to devise an efficient, continuous architecture that retains the scalability of the Mixture of Experts while omitting the top-k operation that limits a token's exposure to different experts. An intuitive way to achieve this is to route all tokens to all experts, but this approach is computationally infeasible for large-scale pretraining. To combat this constraint, the method explored in this work considers what happens not to an individual token but to a whole group of tokens instead. The main contribution of this work is the observation that allowing an expert to dynamically produce a continuous representation of the entire group of tokens that is more lightweight to process than each token individually yields positive results.

More specifically, in our design, an input batch is divided into groups of tokens, and each group is processed independently. Given a group and a single expert, a scalar weight is produced for each token. The weights are then normalized and used to compute a linear combination of the tokens, which is used as the expert's input. The experts' outputs are used for token updates as follows: for each input token, its update is a linear combination of expert outputs, with the token's mixing weights for each expert as coefficients\footnote{The authors note that an MoT layer admits an efficient vectorized implementation, where all meaningful computations are done with batched matrix multiplications.}. A diagram of our method is presented in Figure~\ref{mot}. 

\begin{algorithm}
\caption{Mixture of Tokens layer} \label{alg:mixing}
\begin{algorithmic}[1]
\For {\text{each} $E$ \text{in experts}}:
    \State $\text{weights}_E$ = Softmax(Linear(\text{tokens}))
    \State $\text{mix} = \sum_i \text{token}_{\,i} * \text{weights}_{\,i,E}$
    \State $\text{output}_E = E(\text{mix})$
    \EndFor
\For{each  i}
\For{each  E}    
    \State $\text{update}_i = \sum_E \text{output}_E * \text{weights}_{\,i,E} $

\EndFor
\EndFor

\end{algorithmic}
\end{algorithm}

To see why this method is scalable, it is helpful to examine the relationship between the number of tokens in a group and the number of experts. Essentially, if the two are equal, the total computation done by experts is the same as in the case of top-1 routing. This allows MoT to benefit from the same parameter scaling as seen in MoE, which we confirm empirically in Section~\ref{scalingg}.

\subsection{More mixtures per expert} \label{sec:granularity}
Building on the design described above, we experiment with feeding more than one mixture into each expert. If done without any further modifications, this would mean a linear increase in computation costs for each extra mixture processed. In order to avoid this extra cost, MoT uses more experts, but each expert has a proportionally reduced hidden dimension. This way, each mixture is processed by a small expert, and the layer's total number of parameters, as well as the number of FLOPs used by all experts, stays the same. We find this design to bring consistent improvements as the number of processed mixtures increases.

\subsection{Token Groups in MoT}

\begin{figure}[h]
    \medskip

    \centering
    \includegraphics[width=0.65\linewidth]{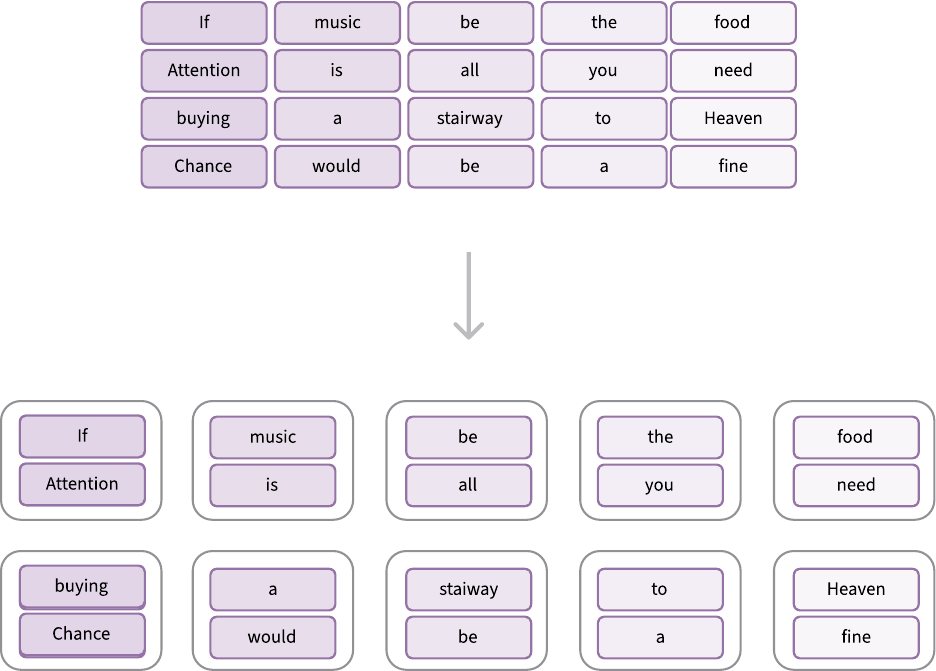}
    \caption{Each group consists of tokens with the same position in a sequence. In this example, the group size is $2$. Note that the maximal possible group size is equal to the batch size.}
    \label{fig:group}
    \medskip
\end{figure}

The question of how token \textit{groups} are decided within a batch is crucial for compatibility with autoregressive training and inference. The key insight is that tokens from the same sequence cannot be placed in one group, as the mixing operation would result in an information leak. Due to this restriction, MoT groups tokens from different examples based on their position in the sequence. Consequently, all tokens within a group have the same position in their respective sequences. As mentioned before, in order to keep the number of FLOPs per token constant, an increase in the number of experts means an equal increase in group size. An illustration of how grouping is done within a batch of tokens is shown in Figure~\ref{fig:group}.

\subsection{Comparison with other Mixture of Experts Architectures}\label{comparison}

\textbf{Scaling.} The technique featured in \citet{hazimeh2021dselectk} is based on a continuously differentiable sparse top-k router, which is a major advantage compared to the common top-k gating. However, this approach requires that all experts are utilized in a portion of training, rendering it computationally prohibitive for models with large numbers of experts. The architecture based on merging experts proposed in \citet{muqeeth2023soft} also presents an attractive, continuous alternative to top-k gating, yet the cost of merging all experts again scales linearly with the number of experts. To combat this, the technique is applied once per sequence, which limits the expressive power of the final model.

\textbf{Training stability.} \citet{lepikhin2020gshard} report instabilities during the training of large MoE models, stemming from the inaccuracies when calculating router weights in low precision. To stabilize the training, they resorted to using full precision. \citet{fedus2022switch} made progress in using mixed precision when training MoE by using selective high precision for gating. When comparing \citet{lepikhin2020gshard,fedus2022switch} to MoT, an advantage of our technique emerges - it is more robust to training in lower precision than other methods. We conjecture, that this is due to the merging mechanism being less susceptible to rounding errors than gating in sparse MoEs. 

\textbf{Token dropping.} Token dropping is a phenomenon where tokens do not receive an update from any expert. This can happen because the expert was chosen by too many tokens in a batch~\citep{fedus2022switch,lepikhin2020gshard,zoph2022stmoe} or, in the case of routing experts to tokens, when a token was not chosen by any expert~\citep{zhou2022mixtureofexperts}. Existing techniques that combat this phenomenon provide a partial solution, but the problem remains. In contrast, tokens in MoT are part of every mixture produced within their group; hence, they always receive an update.

\textbf{Auto-regressive decoding.} Mixture of Tokens is based on the notion of merging tokens prior to being processed by an expert. An encoder-only design of similar nature is featured in concurrent work~\citep{softmoe}. The technique is based on merging patches within an image for vision models. The crucial difference between this technique and MoT is that MoT is compatible with autoregressive training and inference.

\section{Experiments} \label{exps}
The focus of this work is to investigate the efficiency of Tokens on autoregressive language modeling. To measure model quality, we pretrain models for a fixed number of tokens and compare final perplexity in accordance with existing MoE literature~\citep{du2022glam,fedus2022switch}. In all experiments, the models are trained on the C4 dataset~\citep{raffel2023exploring} and use the GPT-2 tokenizer. Unless specified otherwise, we use mixed precision, where all heavy computation is done in bfloat16, whereas the optimizer state and the master weights are kept in full precision. To study the stability of our model, we experiment with training fully in reduced precision.

\begin{figure}[h]
    \centering
    \includegraphics[width=0.65\linewidth]{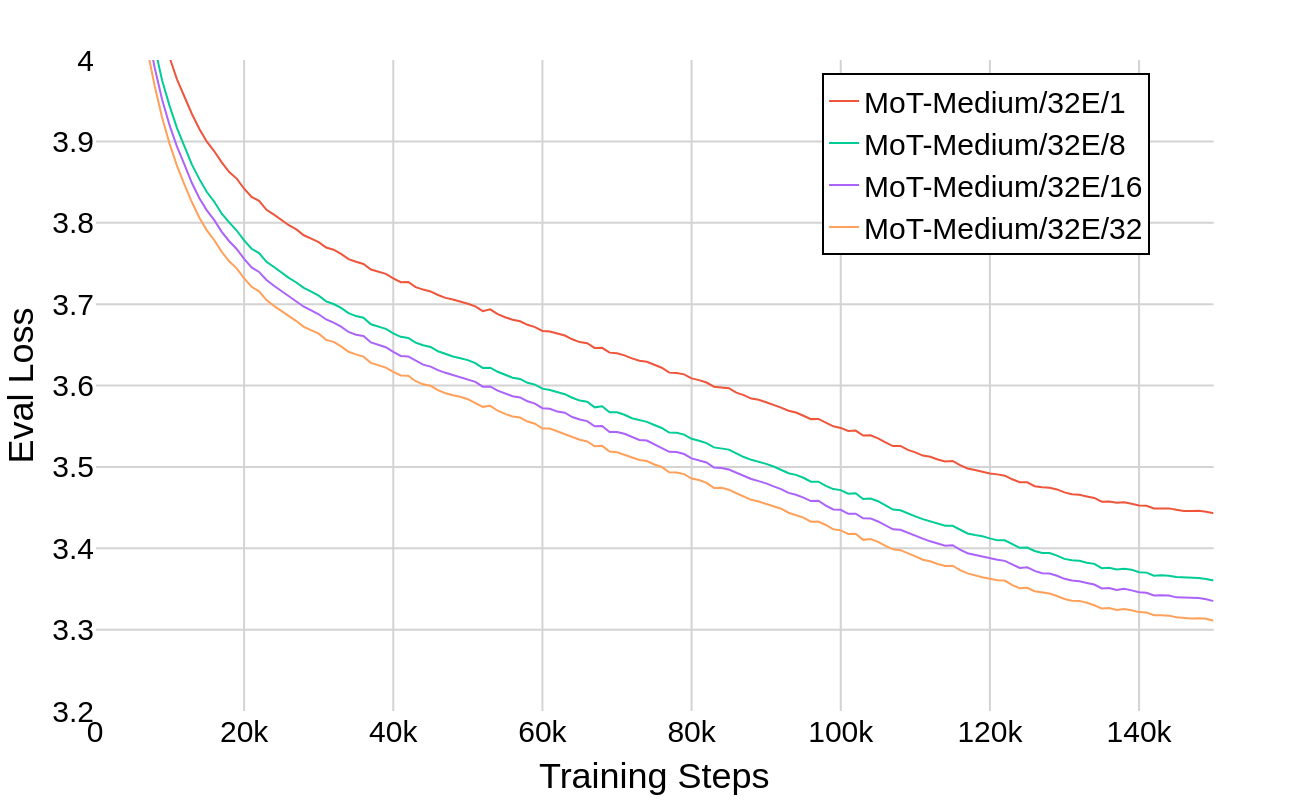}
    \caption{Scaling with respect to the number of token mixtures.}
    \label{fig:granularity}
\end{figure}

Our main result is a substantial speed-up of MoT models compared to dense Transformers (Figure~\ref{fig:main_result}) and results comparable to sparse MoEs (Figure~\ref{fig:vs_smoe}). What follows is the analysis of scaling properties of the MoT architecture with respect to the number of parameters (Figure~\ref{fig:scaling}) and the number of mixtures sent to each expert (Figure~\ref{fig:granularity}). We investigate the model's performance in low precision in order to simulate training instability and find that MoT is less susceptible to instabilities that stem from low-precision training. Lastly, we show the connection between MoT and MoE by spending an additional fraction of pretraining compute to effectively turn a MoT model into a Token Choice model (Section~\ref{trans_tune}).

\subsection{Model Architecture}
The base of our experiments is a decoder-only Transformer based on GPT-2 \citep{gpt2}. We experiment on two model scales: a $77$M Medium model and a $162$M Base model (refer to Appendix~\ref{append1} for hyperparameters and training details). To then obtain a Mixture of Tokens model, we replace the second half of feed-forward layers in the Transformer with MoT layers. Similar to MoE models, the FLOPs and parameter counts in MoT are decoupled. We denote the model architecture by its dense counterpart in terms of the number of FLOPs and, separately, the number of experts (or, equivalently, group size).
To this end, a MoT-Medium/$32$E model is one that uses the same number of FLOPS as a Medium ($77$M) Transformer model but uses $32$ experts in MoT layers.

Regarding the design described in Section~\ref{sec:granularity}, Medium/$32$E/$4$ denotes a model that uses MoT layers with $32 \cdot 4$ small experts that collectively have the same number of parameters as $32$ normal experts.

Besides using the Transformer as a baseline, we also compare against Token Choice~\citep{fedus2022switch} and Expert Choice~\citep{zhou2022mixtureofexperts} as sparse MoE baselines. As Expert Choice is sensitive to the batch size, to avoid discrepancies between training and inference, we group tokens prior to routing in training Expert Choice models.

\subsection{Scaling Results} \label{scalingg}
\begin{figure}[h]
    \centering
    \includegraphics[width=0.65\linewidth]{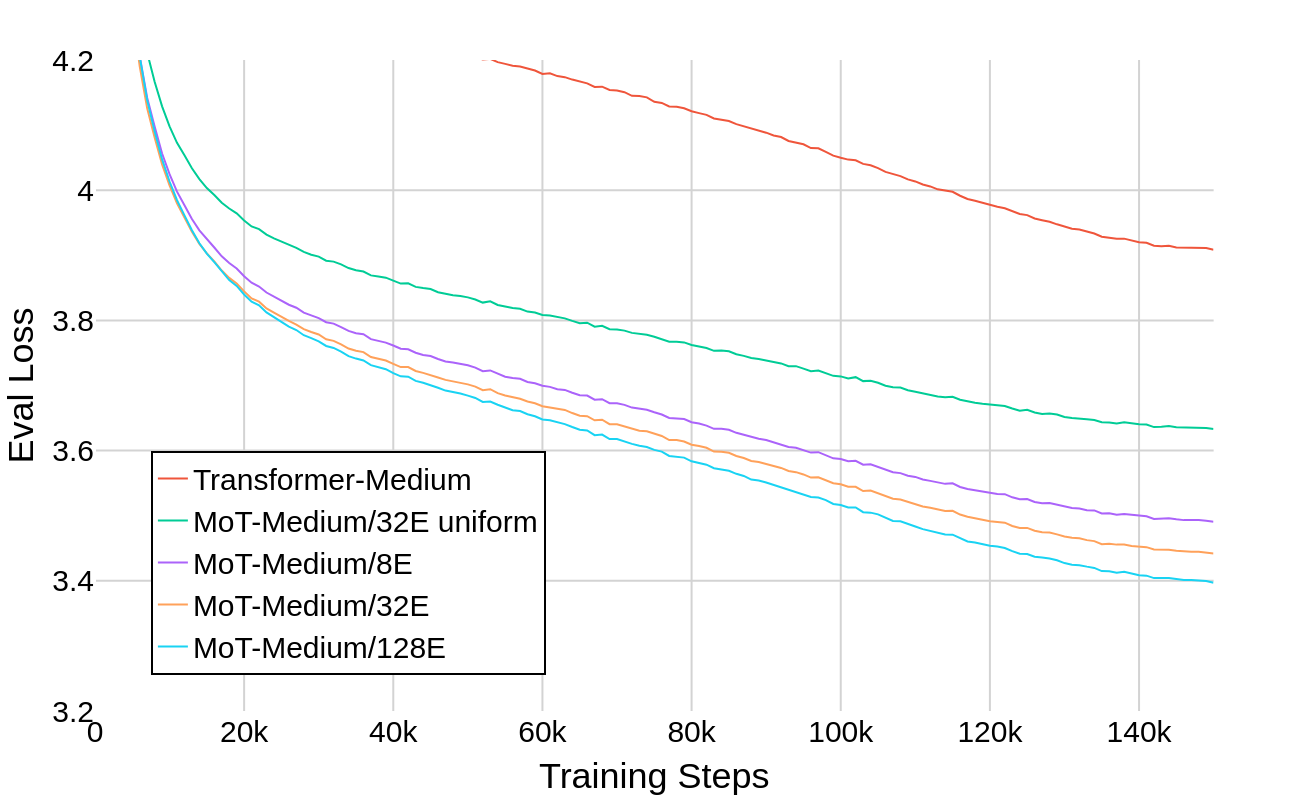}
    \caption{Scaling with respect to the number of parameters. Also featured are the Transformer baseline and an MoT model with a non-learnable, uniform routing strategy.}
    \label{fig:scaling}
\end{figure}

Mixture of Tokens models demonstrate strong scaling properties with respect to the number of parameters. As seen in Figure~\ref{fig:scaling}, increasing the number of experts in MoT layers while using the same compute budget yields consistent improvements. All MoT models are a strict improvement over the Transformer. The figure also features an ablation experiment, where the mixing weights are fixed to $1/n$, with $n$ as the group size. This corresponds to a uniform mixing strategy; the performance of that model clearly suffers, confirming that MoT layers learn non-trivial mixing strategies.

The increased number of token mixtures described in Section~\ref{sec:granularity} is another axis of scaling for MoT models, again exhibiting consistent improvements. We hypothesize that this phenomenon is due to two mechanisms: first, the model is simply more expressive with a larger number of smaller experts, and second, the model can allocate its focus (the mixing weights) more flexibly to more important tokens while reducing the updates for trivial ones.

\subsection{Comparison with the Transformer and Sparse MoEs}

\begin{figure}[h]
    \centering
    \includegraphics[width=0.65\linewidth]{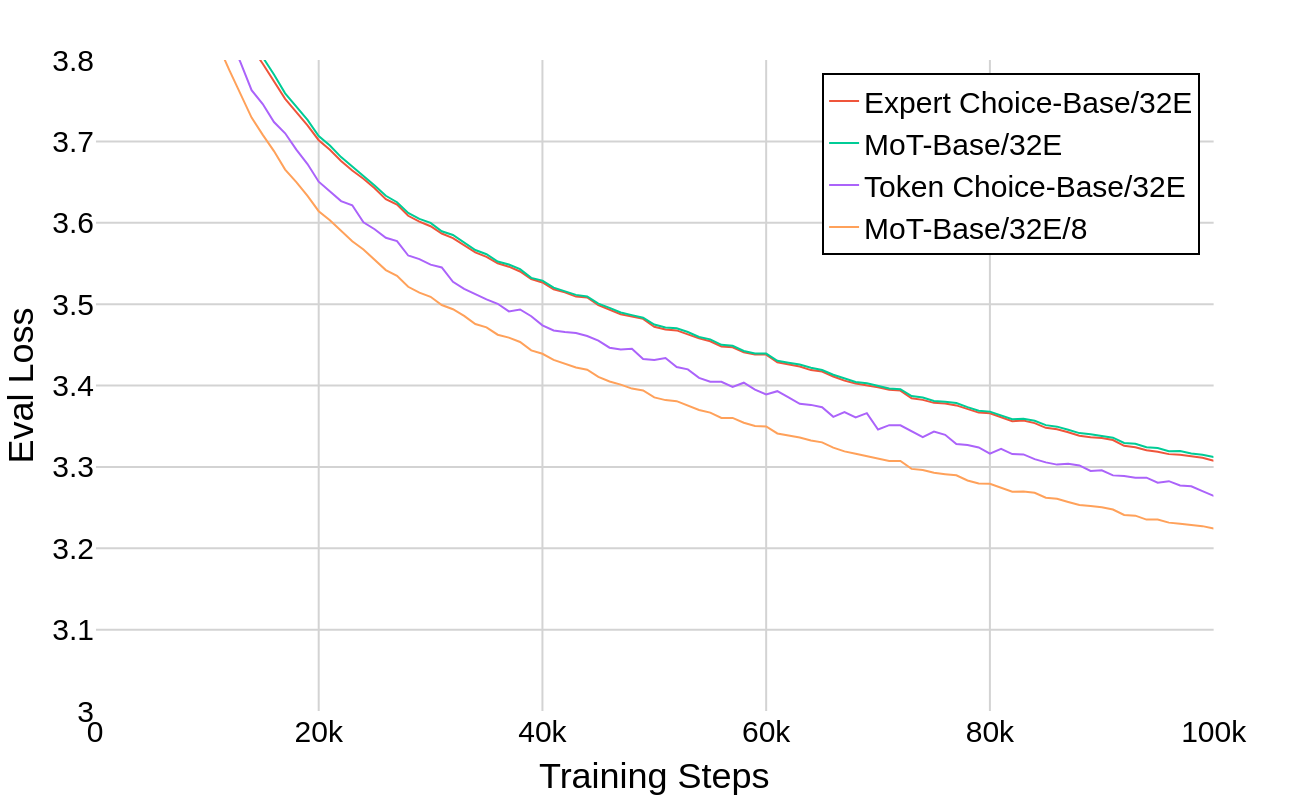}
    \caption{Comparison of MoT and sMoE architectures. An increased number of smaller experts allows MoT to match the performance of the best sMoE model. Due to computational constraints, the models were trained for 100K steps.}
    \label{fig:vs_smoe}
\end{figure}

Crucially, the performance of Mixture of Tokens is comparable to that of the strong Mixture of Experts baselines (Figure~\ref{fig:vs_smoe}). An increased number of mixtures allows it to compete with both Expert Choice and Token Choice architectures. As sparse routing is hypothesized to contribute to training instabilities in large sparse models, Mixture of Tokens, being continuous, presents a promising alternative. To investigate training instabilities at the scale we experiment on, we trained models fully in bfloat16, as opposed to the mixed precision used in all other experiments. The results confirm that MoT is more resistant to lower precision training: as the precision of training decreases, the performance of Expert Choice drops below that of Mixture of Tokens, despite the former attaining better perplexity using mixed precision. We find this to be evidence of the architecture's potential for stable training at higher model scales. See Table~\ref{tab:precision} for details.

\begin{figure}
    \centering
    \includegraphics[width=0.65\linewidth]{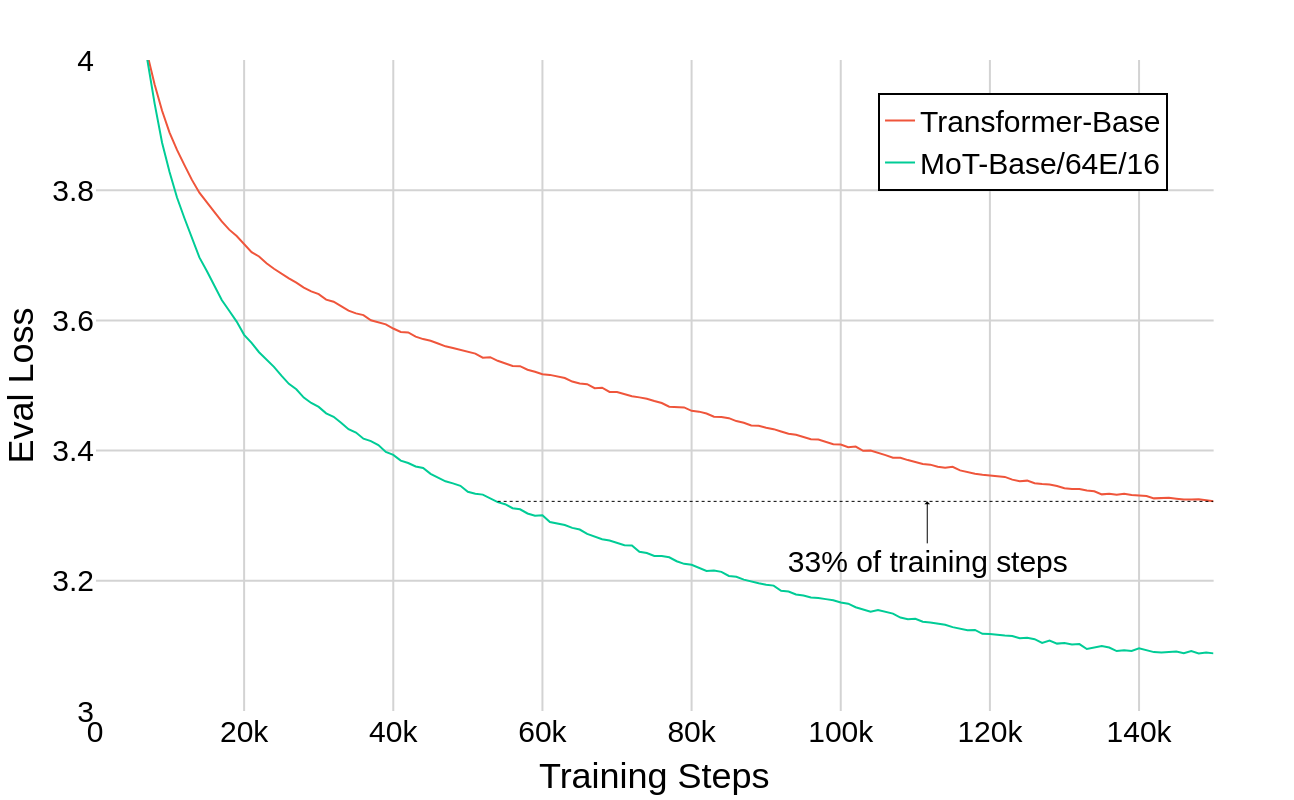}
    \caption{Our best MoT model reaches the final loss of the baseline in just 33$\%$ of the compute budget.}
    \label{fig:main_result}
\end{figure}

Lastly, we combine our findings on MoT scaling properties to train our most efficient MoT model and compare it to the Transformer baseline (Figure~\ref{fig:main_result}). The result is a model that attains the final loss of the baseline in a third of the training steps. This represents a \textbf{$3\times$} improvement in terms of the compute budget.

\begin{figure}[h]
    \centering
    \includegraphics[width=0.65\linewidth]{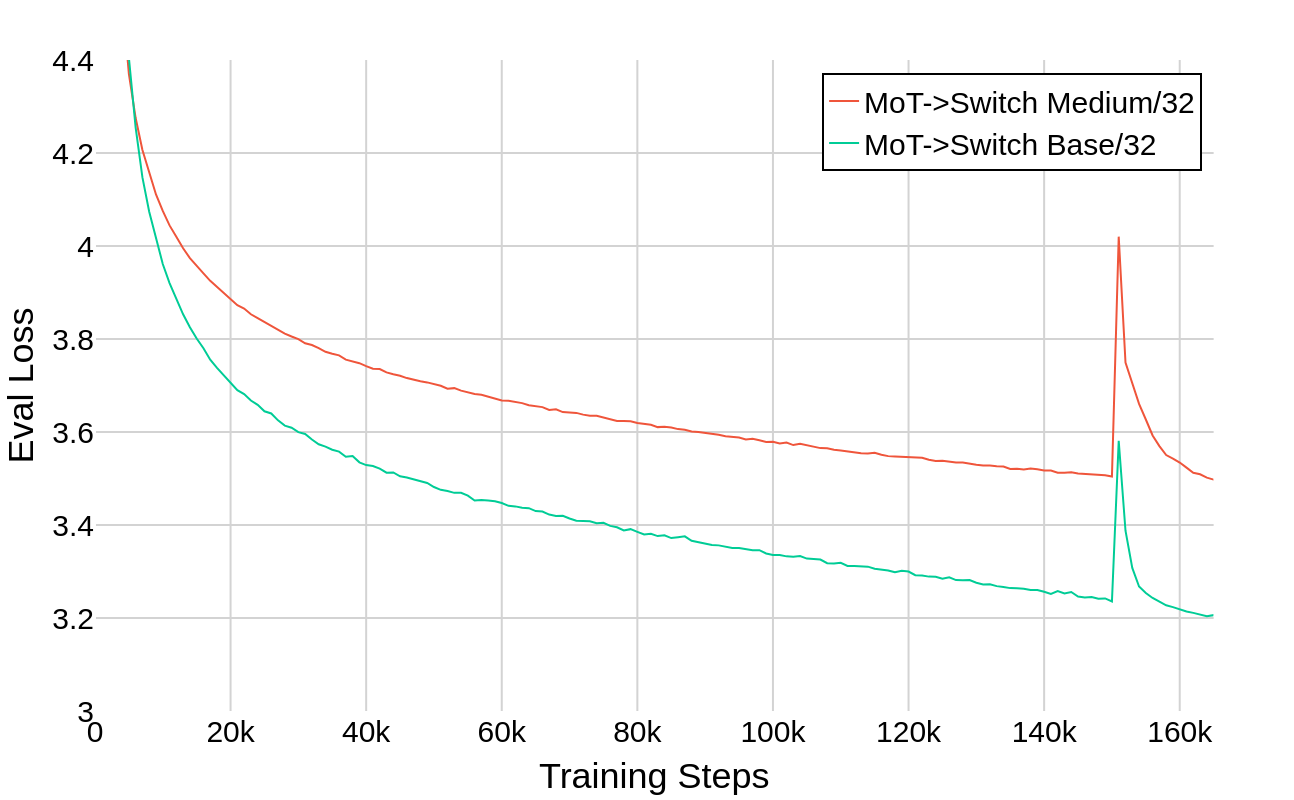}
    \caption{Transition tuning: The first $150$K steps of the model are done using Mixture of Tokens architecture. Then, a new Token Choice model is initialized with weights taken from the MoT model, and the model trains for a further $15$K steps to recover performance. The spike in loss is due to the sudden change of architecture.}
    \label{fig:trans_pic}
\end{figure}

\begin{table}[h]
    \caption{Lower precision training result loss comparison. MoT performs better in the bfloat16-only setting. Learning rates were separately tuned in lower precision for both Expert Choice and MoT. Results averaged over 3 random seeds.}
    \vspace{6pt}
    \label{tab:precision}
    \centering
    \resizebox{0.8\columnwidth}{!}{%
    \begin{tabular}{l  c  c}
        \hline
        & MoT-Medium/$32$E & Expert Choice-Medium/$32$E \\
        \hline
        Mixed Precision & 3.442 ($\pm$ 0.002) & \textbf{3.420 ($\pm$ 0.002)}\\
        bf16 only & \textbf{3.661 ($\pm$ 0.007)} & 3.728 ($\pm$ 0.044) \\
        \hline
    \end{tabular}
    }
\end{table}

\newpage
\subsection{Transition Tuning} \label{trans_tune}
Mixture of Tokens suffers from a drawback common to MoEs, namely, it does not support unbatched inference. This is a direct consequence of its design --- in the forward pass, it groups several tokens from different examples in the batch. With the growing adoption of Large Language Models to consumer hardware~\citep{touvron2023llama,cerisara}, the lack of support could invalidate the architecture's wider adoption. While Mixture of Tokens with a group size of one is technically possible, in order to keep FLOPs constant, the layer would need to reduce trivially to a standard Transformer MLP.

To address this, we demonstrate that the weights learned by Mixture of Tokens can be used to directly initialize a Token Choice model with the same specifications (number of experts and expert size). The layer responsible for producing mixing weights is used to initialize the sparse router. To mitigate the performance difference caused by this change of architecture, we train the entire new model (no weights are frozen) for $10\%$ of the total pretraining steps of the original model in order to recover the original model's performance (measured in evaluation loss). We call this technique \textit{transition tuning}. This way, it is possible to train with Mixture of Tokens and enjoy unbatched generation at inference time. We hypothesize that this pipeline would be especially attractive in setups where parts of the model cannot be trained in higher precision, such as on specialized, low-precision hardware. The results are presented in Figure~\ref{fig:trans_pic}.

\section{Limitations and Future Work}
\label{limitations}
With the strong performance of MoT on medium-sized models, an obvious next step is to train larger models. This presents an opportunity to validate stability results on larger models, where training instabilities are more common.

As with most Mixture of Experts models, the memory footprint of MoT layers is substantial. Scaled models require large amounts of RAM on specialized hardware for training, making their adoption expensive. To this end, an attractive future direction would be to investigate model distillation with Mixture of Tokens models.

Lastly, both training and inference with MoT mix different examples within a single batch. This mixing of tokens from different sequences and the necessity of performing batched inference may be undesirable in some use cases. While performing unbatched inference is always inefficient with LLMs, as the memory throughput to access model weights becomes the bottleneck, unbatched inference still finds its uses. While transition tuning solves this problem, exploring different inference strategies might bring new insights.

\section{Conclusions}

In this work, we presented the Mixture of Tokens, a novel continuous Mixture of Experts architecture compatible with autoregressive decoding. This architecture scales to model sizes similar to sparse Mixture of Experts, matches its performance, and is more resistant to training instabilities due to lower precision training. Moreover, we introduced transition tuning, a technique for initializing an MoE model with another pretrained MoE model of a different architecture, and showed that the new model attains the performance of the original one using a fraction of the compute budget.

\section*{Acknowledgements}

We would like to express sincere gratitude to Piotr Padlewski and Tomasz Trzciński for general feedback and Dagmara Rudzińska for invaluable support with graphic design.

This work was funded by IDEAS NCBR, which also provided significant computational resources. Marek Cygan was partially supported by an NCBiR grant POIR.01.01.01-00-0392/17-00. The research was supported by PL-Grid infrastructure (grant PLG/2023/016148). We also benefited from the Entropy cluster (hosted at the Faculty of Mathematics, Informatics and Mechanics of the University of Warsaw) funded by NVIDIA, Intel, the Polish National Science Center grants UMO-2017/26/E/ST6/00622 and 2022/45/N/ST6/02222, and ERC Starting Grant TOTAL.

\bibliographystyle{unsrtnat}  
\bibliography{references}

\newpage
\appendix
\section{Training Hyperparameters} \label{append1}

All models were trained using mixed precision unless explicitly stated. We trained all experiments with a batch size of $256$ and a context length of $256$ for $150K$ training steps (unless explicitly stated), resulting in a total of $10$B training tokens. We used the AdamW optimizer with default hyperparameters. When required, we employed a Fully Sharded Data Parallel approach from PyTorch to parallelize the training across multiple machines. Learning rates were tuned separately depending on model size and architecture. The optimal learning rate for Transformers was 1e-3 for Medium and 4e-4 for Base models, while for both MoT and MoE, they were 7e-4 for Medium and 2e-4 for Base. 
\begin{table}[h]
    \vspace{10pt}
    \centering
    \caption{Training hyperparameters. The table provides example models featured in experiments. All remaining models can be derived from this table.}
    \resizebox{\columnwidth}{!}{%
    \begin{tabular}{cccccccccc}
        \toprule
        Model & Experts & Expert & Group & Total & Blocks & $d_{model}$ & $d_{ff}$ & \#att. \\
         & & size & size & params & &  & &heads \\
        \midrule
Transformer-Medium & - & - & - & 77M & 8 & 512 & 2048  & 8  \\
MoT-Medium/32E & 32 & 2048 & 32 & 336M & 8 & 512 & -  & 8  \\
MoT-Medium/32E/8 & 256 & 256 & 32 & 337M & 8 & 512 & -  & 8 \\
Transformer-Base & - & - & - & 162M & 12 & 768 & 3072 & 12 \\
MoT-Base/32E & 32 & 3072  & 32 & 520M & 12 & 768 & - & 12 \\      
MoT-Base/64E/16 & 1024 & 192 & 64 & 977M & 12 & 768 & - & 12 \\      
        \bottomrule
    \end{tabular}
    }
    \label{tab:hyperparams}
\end{table}  

\section{Reproducibility} 

The code and configuration files used to produce the results described in this work are available in our public repository at \url{https://github.com/llm-random/llm-random}.

\section{Contributions}
\label{sec:contributions}

Szymon implemented a PoC and different variants of MoT, together with running experiments and optimizations. Michał implemented and experimented with various MoT designs and contributed to the infrastructure design and implementation. Sebastian provided the initial idea, research intuitions, and direct project supervision. Maciej was responsible for parts of evaluation and significant engineering. Jakub implemented MoE baselines, Jan stabilized Mixture of Experts training, while both helped with MoE hyperparameter tuning. Tomasz consulted ideas and helped with cluster infrastructure. Kamil and Krystian contributed to general engineering. Everybody above contributed to the infrastructure of the project. Marek provided scientific advice and high-level supervision.

\end{document}